\documentclass[10pt,twocolumn,letterpaper]{article}

\usepackage[pagenumbers]{cvpr}

\usepackage[dvipsnames]{xcolor}

\usepackage{multirow}
\definecolor{cvprblue}{rgb}{0.21,0.49,0.74}
\usepackage[pagebackref,breaklinks,colorlinks,citecolor=cvprblue]{hyperref}
\usepackage{graphicx}
\usepackage{amsmath}
\usepackage{amssymb}
\usepackage{booktabs}
\usepackage{multirow}
\usepackage{svg}
\usepackage[accsupp]{axessibility}

\newcommand{\vect}[1]{\mathbf{#1}}

\newcommand{\mat}[1]{\mathbf{#1}}
\newcommand{\normal}{\mathcal{N}}

\newcommand{\norm}[1]{\lVert #1 \rVert}
\newcommand{\argmin}[1]{\underset{#1}{\arg\min}}

\title{Learned representation-guided diffusion models for large-image generation}

\author{
Alexandros Graikos\thanks{Equal contribution. Correspondence to \href{mailto:agraikos@cs.stonybrook.edu}{agraikos@cs.stonybrook.edu}}\ 
\quad Srikar Yellapragada\textsuperscript{*} \quad Minh-Quan Le 
\\ \quad Saarthak Kapse \quad Prateek Prasanna \quad Joel Saltz \quad Dimitris Samaras \\
Stony Brook University
}

\definecolor{orange}{HTML}{FF7E00}
\definecolor{teal}{HTML}{48D1CC}

\begin{document}
\maketitle

\begin{abstract}
To synthesize high-fidelity samples, diffusion models typically require auxiliary data to guide the generation process. However, it is impractical to procure the painstaking patch-level annotation effort required in specialized domains like histopathology and satellite imagery; it is often performed by domain experts and involves hundreds of millions of patches. Modern-day self-supervised learning (SSL) representations encode rich semantic and visual information. In this paper, we posit that such representations are expressive enough to act as proxies to fine-grained human labels. We introduce a novel approach that trains diffusion models conditioned on embeddings from SSL. Our diffusion models successfully project these features back to high-quality histopathology and remote sensing images. In addition, we construct larger images by assembling spatially consistent patches inferred from SSL embeddings, preserving long-range dependencies. Augmenting real data by generating variations of real images improves downstream classifier accuracy for patch-level and larger, image-scale classification tasks. Our models are effective even on datasets not encountered during training, demonstrating their robustness and generalizability. Generating images from learned embeddings is agnostic to the source of the embeddings. The SSL embeddings used to generate a large image can either be extracted from a reference image, or sampled from an auxiliary model conditioned on any related modality (e.g. class labels, text, genomic data). As proof of concept, we introduce the text-to-large image synthesis paradigm where we successfully synthesize large pathology and satellite images out of text descriptions.
\footnote{Code is available at \href{https://github.com/cvlab-stonybrook/Large-Image-Diffusion}{this link}} 
\end{abstract}

\section{Introduction}
\label{sec:intro}

\begin{figure}[ht]
    \centering
    \includegraphics[width=\linewidth]{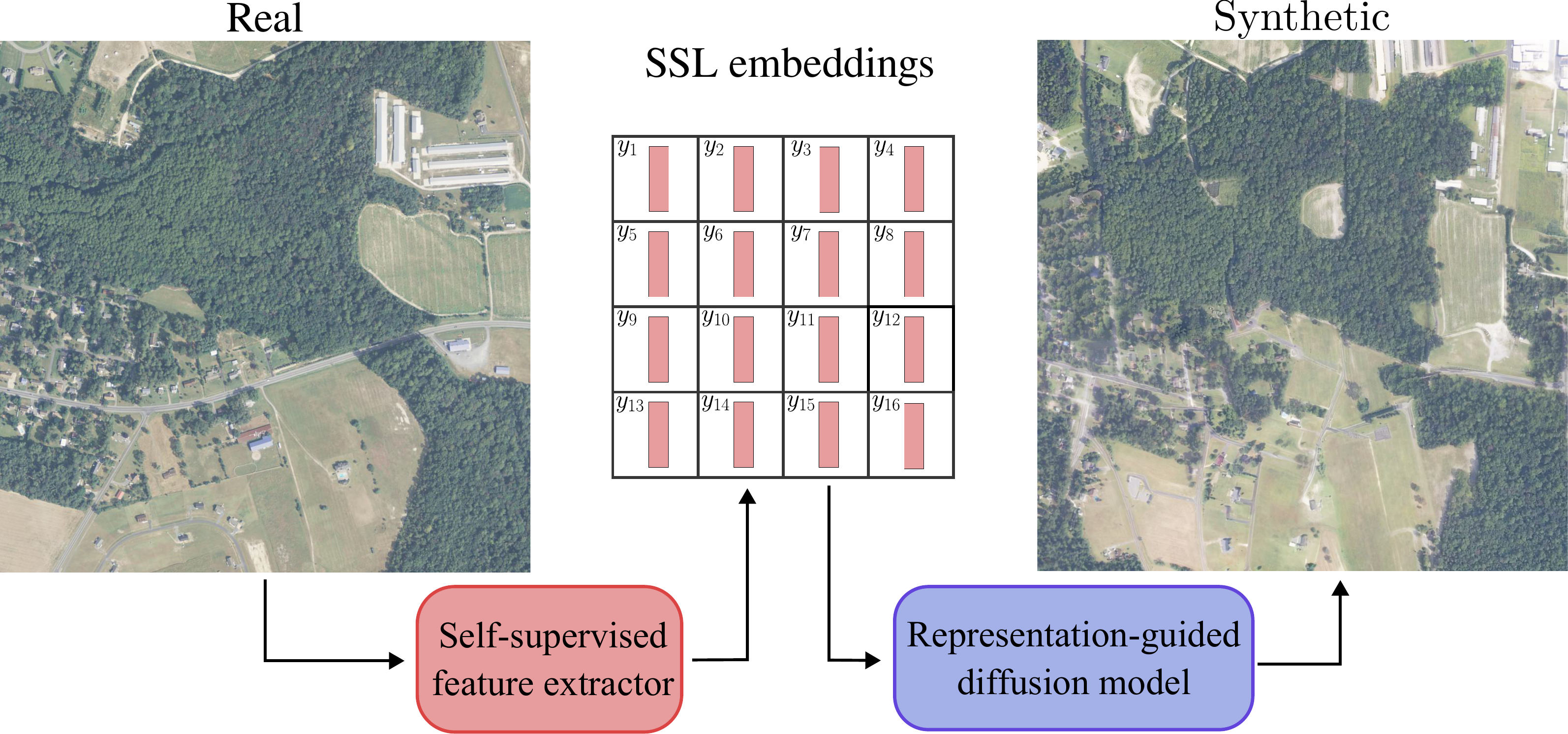}
    \caption{We propose using SSL features to condition diffusion models. This allows us to construct large images by assembling consistent patches inferred from a spatial arrangement of SSL embeddings. The generated image retains the semantics of the embeddings used as a condition, maintaining the forested and open areas from the reference. Best viewed zoomed-in.}
    \label{fig:teaser}
\end{figure}

Diffusion models produce high-quality and diverse samples across a spectrum of generative tasks~\cite{dhariwal2021diffusion, liu2023audioldm}. This leap forward has been enabled by the simultaneous curation of large-scale multi-modal datasets~\cite{schuhmann2022laion} and the development of efficient conditioning mechanisms~\cite{ramesh2022hierarchical,rombach2022high}. The key to unlocking the models' capabilities is to integrate auxiliary information during training and inference~\cite{nichol2021improved, dhariwal2021diffusion, ho2022classifier}. 

Large-scale human-annotated datasets are mostly limited to image-caption pairs, collected from easily accessible online repositories and labeled by non-expert annotators. However, in domains such as digital histopathology and remote sensing, where gigapixel scale images provide vast amounts of unlabeled data, annotation proves challenging. Moreover, the process requires expert knowledge and is more difficult at a finer scale, i.e., captioning large crops of the gigapixel image is simpler than captioning smaller patches. Based on our estimates (see supplemental), annotating the entire TCGA-BRCA dataset with captions would take $\approx$40,000 hours of pathologist's time. Replicating the impressive results of diffusion models in these domains has been limited by the scarcity of fine-grained per-image conditioning, vital for high-quality image synthesis~\cite{nichol2022glide}.

Modern-day self-supervised learning (SSL) representations~\cite{caron2021emerging} encode rich semantic and visual information. Features from trained self-supervised models serve as compact image representations and are widely used to perform discriminative downstream tasks successfully~\cite{chen2022scaling, wang2022transformer, filiot2023scaling}, proving that these compressed representations indeed encode useful semantic information about the images. We hypothesize that such SSL representations are already expressive enough to act as proxies to fine-grained human labels. If this is true, these representations should be able to condition the training of effective diffusion models in these domains. In this novel approach, we utilize self-supervised feature extractors as image annotators; these features provide necessary per-image conditioning signals at the highest resolutions, required for diffusion model training.

Our experiments show that conditioning with expressive self-supervised features leads to precise control over the image content. The SSL features are adept at identifying complex patterns and structures in the images, while the diffusion model learns to translate them into visual components accurately (Fig.~\ref{fig:real_synthetic}). This motivates us to perform large-image synthesis by locally controlling the appearance using SSL conditioning and dictating the global structure through the spatial arrangement of the conditions.

Our approach synthesizes large images in a patch-based manner, using a single image diffusion model at the highest resolution. We represent a large image as a grid of SSL embeddings, where each serves as a representation of a large image neighborhood. The whole image is then synthesized by generating consistent patches that capture both the local properties, as given by the local patch conditioning, and the spatial arrangement of the conditioning features. If we change this spatial arrangement, we are, in fact, editing how semantic elements are arranged globally in the large image. This strategy enables the controllable generation of images of virtually any size without significantly increased computation compared to the base patch-level model.

To generate a large image, our approach requires the patch diffusion model and the spatially-arranged conditioning. We can start with a reference large image as source and extract SSL embeddings from non-overlapping segments, enabling our method to synthesize a variation of the original image (Fig.~\ref{fig:teaser}). 

Utilizing SSL embeddings as conditions allows us to have the necessary control over image generation, at the expense of an explainable and easy-to-use conditioning mechanism. Nevertheless, we argue that since generating images from learned embeddings is agnostic to the embedding source, there are simple ways to combine control over generated images with explainability. We propose training auxiliary models to transform higher-level conditioning signals, such as text captions, to the learned patch representations. To demonstrate this versatility, we introduce text-to-large-image synthesis by training an auxiliary model to sample a spatial arrangement of embeddings from a text description.

We train patch-level diffusion models using self-supervised features as conditioning on digital histopathology (TCGA~\cite{cancer2013cancer}) and satellite image (NAIP~\cite{naip}) datasets. We perform extensive evaluations and demonstrate the advantages of SSL conditioning and our large-image generation framework on synthesis and classification tasks. Our model achieves exceptional patch-level and large-image quality, the ability to improve classifiers through data augmentation even when synthesizing out-of-distribution data, and effective fusion of diffusion and SSL features for downstream applications. Finally, we are the first to perform text-to-large image synthesis, which should be of significant community interest as vision-language models (VLMs) for pathology and satellite images gain traction.

In summary, our contributions are as follows: 
\begin{itemize}
    \item We develop a novel method to train diffusion models with self-supervised learning features as conditioning and generate high-quality images in the histopathology and satellite image domains.
    \item We present a framework for large-image synthesis, based on self-supervised guided diffusion, that maintains contextual integrity and image realism over large areas. 
    \item We demonstrate the applicability of our model in various classification tasks and showcase its unique ability to augment out-of-distribution datasets. 
    \item We introduce text-to-large image generation for digital histopathology and satellite images, highlighting the versatility of our approach.
\end{itemize}

\section{Related work}
\label{sec:related}
\textbf{Diffusion models:} Introduced for image generation by Ho \etal~\cite{ho2020denoising}, diffusion models have evolved considerably. These enhancements include class conditioning~\cite{nichol2021improved}, architectural improvements and gradient-based guidance~\cite{dhariwal2021diffusion}, and classifier-free guidance~\cite{ho2022classifier}. Latent Diffusion Models (LDMs)~\cite{rombach2022high} proposed a two-step training process with a Variational Autoencoder (VAE) compressing input images into a lower-dimensional latent space and a diffusion model trained in this latent space. Denoising Diffusion Implicit Models (DDIM)~\cite{song2020denoising} accelerate the sampling process by $10-50 \times$. Self-guided diffusion models~\cite{hu2023self} also utilize self-supervised learning by quantizing SSL embeddings. In contrast to our approach, their quantization discards useful information from the SSL embeddings.

Training generative models directly at the gigapixel resolution is infeasible. An alternative is to generate the images in a coarse-to-fine manner hierarchically. This has already been applied to natural images, where chaining multiple diffusion models generates images up to $1024 \times 1024$ resolution~\cite{podell2023sdxl, saharia2022image}. However, it is still limited, as it substantially increases the parameter count and is inherently constrained by the final target resolution. 

\begin{figure*}[ht]
    \centering
    \includegraphics[width=.9\linewidth]{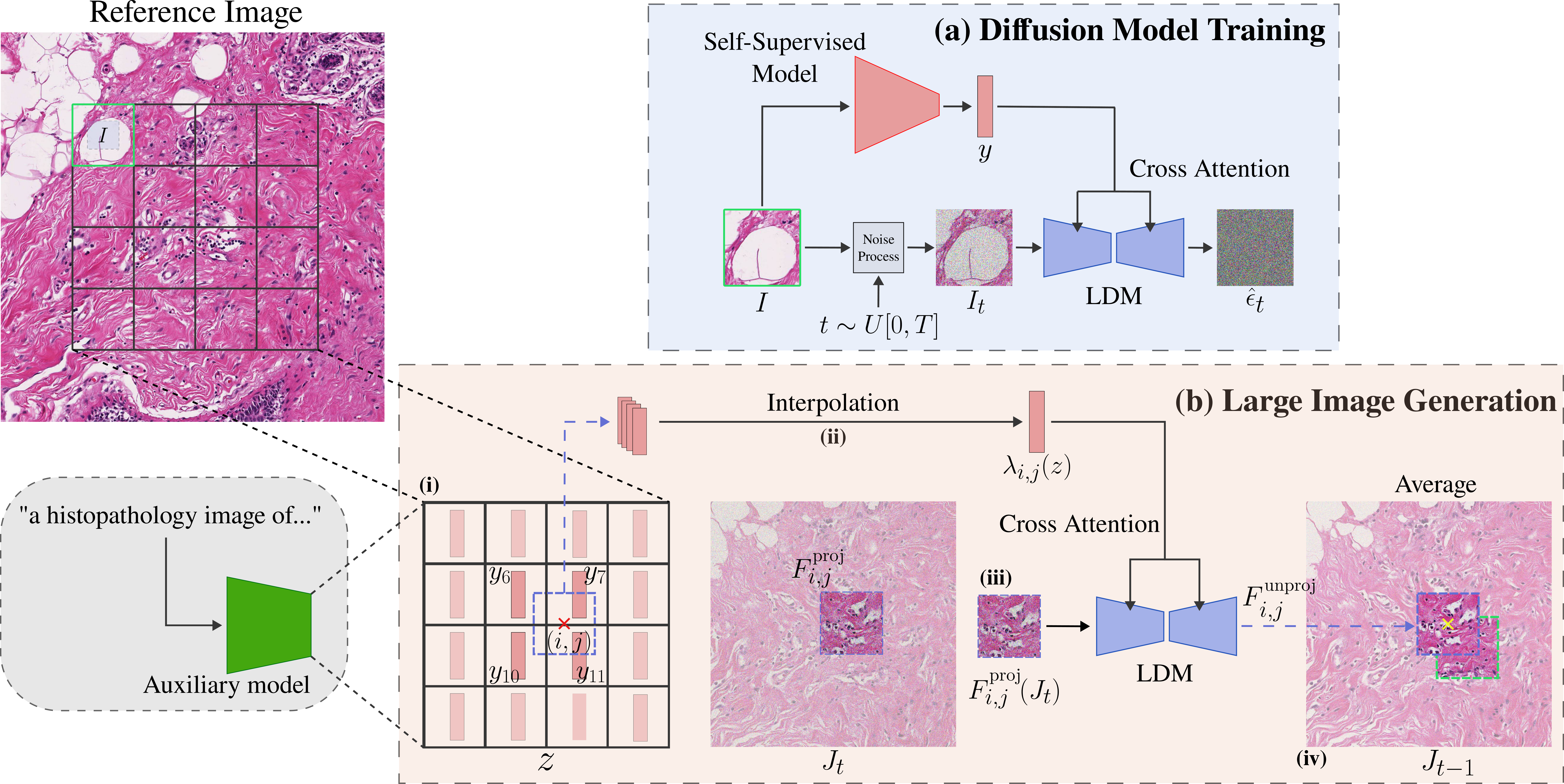}
    \caption{(a) We train diffusion models on patches $I$ (e.g. the one in the green box) taken from a large image conditioned on SSL embeddings. (b) We present our large image generation framework in 4 steps: (i) We extract a set of spatially arranged embeddings from a reference image or sample them from an auxiliary model. (ii) For every location $(i,j)$, we compute a conditioning vector $\lambda_{i,j}$ by interpolating the spatial grid of embeddings. (iii) At every diffusion step, we denoise the patch $F(i,j)$ using the conditioning $\lambda_{i,j}$. (iv) The next step is computed by averaging the denoising updates of all patches that overlap at $(i,j)$.}
    \label{fig:method}
\end{figure*}

In the context of digital histopathology, works have been limited to training unconditional~\cite{moghadam2023morphology} or class-conditioned~\cite{xu2023vit,muller2023multimodal} diffusion models. For text conditioning, pathology text reports were used to provide context on the whole-slide scale~\cite{yellapragada2023pathldm}. Similar approaches have been applied to satellite data~\cite{espinosa2023generate, sebaq2023rsdiff}. Apart from high-quality images without manual annotation, our SSL conditioning is also necessary for large-image synthesis, as all previous conditioning methods would not be sensitive to the intricate differences between neighboring patches. In recent work, DiffInfinite~\cite{aversa2023diffinfinite} explored large-image generation using segmentation masks as conditioning. We argue this is still sub-optimal as it requires accurate, human-annotated masks for training and a mask-generating model during inference.

\textbf{Self-Supervised Learning} Self-supervised learning (SSL) refers to both discriminative~\cite{he2020momentum, bao2021beit} and distillation~\cite{caron2021emerging} approaches that aim to learn representations of the data without supervision. In this work, we are mainly interested in self-supervised learning for histopathology. Notable recent developments include the Hierarchical Image Pyramid Transformer (HIPT)~\cite{chen2022scaling}, which utilizes the inherent hierarchical structure of Whole Slide Images (WSIs), CTransPath~\cite{wang2022transformer}, a hybrid CNN and multi-scale Swin Transformer model, and iBOT~\cite{filiot2023scaling}, a masked image modeling method. These models excel as patch-level feature extractors for WSIs by leveraging the unique structure of the large-image data and capture the important semantic information that we require for conditioning a generative model.

\section{Method}
\label{sec:method}
We propose training a diffusion model on patches $I$ drawn from large histopathology and satellite images, using self-supervised embeddings $y$ as conditioning. Furthermore, we present a patch-based approach that utilizes the SSL-conditioned diffusion model to synthesize arbitrarily large images. An overview of our method is presented in Fig.~\ref{fig:method}.

\subsection{Learned representation-guided diffusion}
\label{subsec:patch_gen}
Given a pre-trained self-supervised feature extractor, we employ LDMs~\cite{rombach2022high} to learn the distribution over the large-image patches $p(\mathcal{I})$. LDMs are comprised of three components: an image-compressing Variational Autoencoder (VAE) that transforms the input images to latent representations, a U-Net denoiser to learn a denoising diffusion process that transforms Gaussian noise to latents, and a conditioning mechanism that controls the diffusion process. The conditioning is performed in our setting with a single vector $y$, obtained from the self-supervised model, integrated via a cross-attention mechanism. We train the LDM using pairs of image patches $I$ and the corresponding extracted self-supervised embeddings $y$.

\subsection{Large image generation}
\label{subsec:large_image_gen}
Our goal is to synthesize high-quality large images, that not only capture global structure but also maintain spatial consistency. As shown in Fig.~\ref{fig:method}, we replicate the semantics in each patch with SSL-guided LDM. At the same time, we preserve the global arrangement of these semantics as defined by the grid of patches in the reference image. Future work can explore alternative approaches to ensure spatial alignment, including using topological constraints or integrating low-resolution information. 

We follow the MultiDiffusion~\cite{bar-tal2023multidiffusion} methodology to generate large images using only a patch-based diffusion model. We can represent the diffusion model as a learned mapping from images and conditions to images:
\begin{equation}
    \Phi : \mathcal{I} \times \mathcal{Y} \rightarrow \mathcal{I},
    \label{eq:diffusion_patch}
\end{equation}
where $\mathcal{I} = \mathbb{R}^{H \times W \times C}$ are the large-image patches and $\mathcal{Y} = \mathbb{R}^{d}$ are the per-patch conditions. To generate a patch, we initialize $I_T \sim \normal(\vect{0}, \mat{I})$ and sequentially transform to a ``clean'' image $I_0$ following the trained LDM:
\begin{equation}
    I_{t-1} = \Phi(I_t \mid y)
\end{equation}
for $t=T,\ T-1,\ \dots,\ 1$. We assume that a large image can be formed as a spatial grid of $P \times P$ patches of size $H \times W$. We then define the large-image diffusion model as
\begin{equation}
    \Psi : \mathcal{J} \times \mathcal{Z} \rightarrow \mathcal{J}
    \label{eq:diffusion_large_image}
\end{equation}
where $\mathcal{J} = \mathbb{R}^{PH \times PW \times C}$ are the large images and $\mathcal{Z} = \underbrace{\mathcal{Y} \times \mathcal{Y} \dots \times \mathcal{Y}}_{P^2}$ are the conditioning vectors of all patches. The process $\Psi(J_t \mid z) = J_{t-1}$, with $J_T \sim \normal(\vect{0}, \mat{I})$, can be approximated by first defining mappings between the two different image and conditioning spaces: 
\begin{equation}
    F^{\text{proj}}_{i,j}  : \mathcal{J} \rightarrow I,\quad \lambda_{i,j} : \mathcal{Z} \rightarrow \mathcal{Y}
    \label{eq:mappings}
\end{equation}
where, for the case of large-image generation, we set $F^{\text{proj}}_{i,j}$ to be a crop (projection) of the large image, centered at $i,j$, and $\lambda_{i,j}$ the conditioning $y_{i,j} \in \mathbb{R}^d$ at $i,j$. Since we only have conditioning vectors at the centers of the patches, we can use spatial interpolation algorithms to implement $\lambda$. This assumes that the interpolant $y_{i,j}$ is a valid conditioning vector for the diffusion process, which we validate experimentally. Then, $\Psi$ is defined as
\begin{equation}
    \Psi(J_t \mid z) = \argmin{J \in \mathcal{J}} \sum_{i,j} \norm{F^{\text{proj}}_{i,j}(J) - \Phi(F^{\text{proj}}_{i,j}(J_t) \mid \lambda_{i,j}(z))}^2
    \label{eq:ftd_minimize}
\end{equation}
which can be solved in closed-form by setting each pixel $i,j$ of $J$ to the average of all the patch-diffusion updates
\begin{equation}
    \Psi(J_t \mid z) = \sum_{i,j} \frac{F^{\text{unproj}}_{i,j}(\vect{1})}{\sum\limits_{k,l} F^{\text{unproj}}_{k,l}(\vect{1})} 
    \otimes F^{\text{unproj}}_{i,j}(\Phi(F^{\text{proj}}_{i,j}(J_t) \mid \lambda_{i,j}(z)))
    \label{eq:diffusion_update}
\end{equation}
where $\vect{1}$ denotes a patch image where all values are set to 1 and $F^{\text{unproj}}_{i,j}$ is the inverse mapping of pixels from the crop centered at $i,j$ back to the large image.

We are able to generate images larger than the ones produced by the patch diffusion model. At the same time, we control what each patch looks like, which is crucial in maintaining the semantic integrity of the larger image. The self-supervised conditions can capture the variations between neighboring patches necessary for producing realistic results. Generating large images with coarser conditioning, such as global text prompts~\cite{yellapragada2023pathldm}, would lead to uniform texture regions (see supplementary). 

\subsection{Controllable large-image synthesis}
\label{sec:controllable_synth}
Although the grid of self-supervised embeddings $z$ cannot be manipulated in a human-interpretable manner, we argue that it is simple to assert more control over the generated images. As illustrated in Fig.~\ref{fig:method}, this control can be attained by training auxiliary models $p(z \mid c)$ that translate higher-level conditioning signals, such as text captions $c$, to learned patch representations $z$.

Since there is no available dataset of paired large images and captions, we resort to pre-trained multi-modal models, such as Quilt~\cite{ikezogwo2023quilt}, CLIP~\cite{radford2021learning} and BLIP~\cite{li2022blip}, to provide the text conditioning. We construct training sets by extracting self-supervised embeddings from training-set large images and pairing them with multi-modal image embeddings or generated captions. During inference, we first sample SSL embeddings from the learned distribution $p(z \mid c)$, then utilize our patch diffusion models to synthesize a large image.

\section{Image Generation Experiments}
\label{sec:image_gen_Exp}

\subsection{Datasets}
We train diffusion models on digital histopathology images from The Genome Cancer Atlas (TCGA)~\cite{cancer2013cancer} and satellite imagery from the National Agriculture Imagery Program  (NAIP)~\cite{naip}. Specifically, we used the TCGA-BRCA (Breast Invasive Carcinoma Collection), TCGA-CRC (COAD + READ Colorectal Carcinoma) datasets, and the Chesapeake Land Cover dataset~\cite{robinson2019large}. 

For the TCGA-BRCA and TCGA-CRC datasets, we use images at $20\times$ magnification, and developed diffusion models conditioned on embeddings from HIPT~\cite{chen2022scaling} and iBOT~\cite{filiot2023scaling}, which were pre-trained on PanCancer TCGA. In the case of the HIPT model, we specifically used its patch-level ViT. Additionally, for the TCGA-BRCA dataset, we train a model at $5\times$ magnification using embeddings from CTransPath~\cite{wang2021transpath} for additional evaluations.

\begin{figure*}[ht]
    \centering
    \includegraphics[width=.9\linewidth]{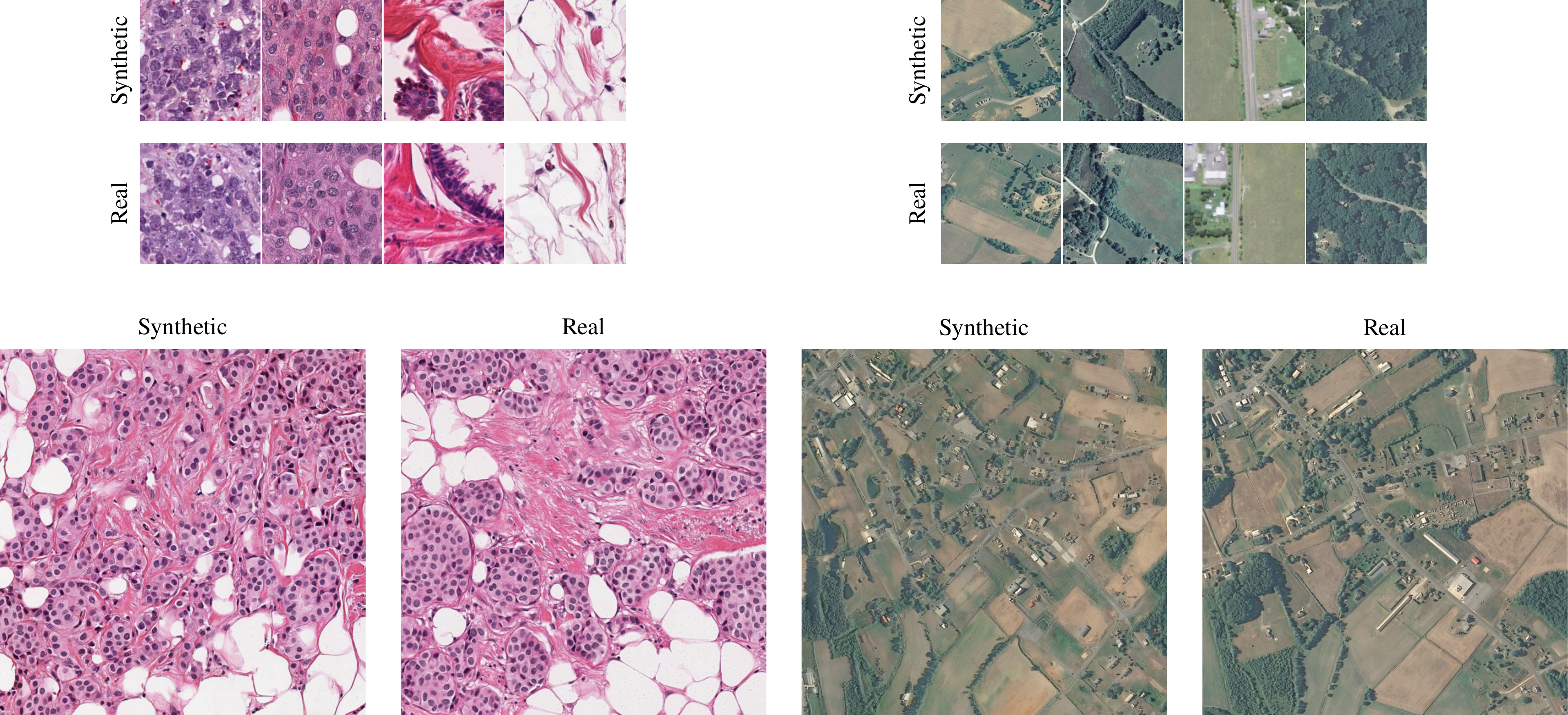}
    \caption{(Top) Patches (256 $\times$ 256) from our models, and the corresponding reference real patches used to generate them. The SSL-guided LDM replicates the semantics of the reference patch.
    (Bottom) Large images (1024 $\times$ 1024) from our models, and the corresponding reference real images used to generate them. We preserve the global arrangement of the semantics defined in the reference image.}
    \label{fig:real_synthetic}
\end{figure*}

The Chesapeake Land Cover dataset dataset contains 732 NAIP tiles, each measuring a 6km $\times$ 7.5 km area at 1m resolution. We extract $256 \times 256$ non-overlapping pixel patches, resulting in 667,000 patches total. Given the absence of publicly available self-supervised learning models tailored to the NAIP data, we train a Vision Transformer (ViT-B/16)~\cite{dosovitskiy2020image} using the DINO framework~\cite{caron2021emerging}. We then use the learned DINO embeddings to train the diffusion model on pairs of image patches and self-supervised embeddings. 

For patch-level augmentation, we employ the NCT-CRC dataset~\cite{kather_2018_1214456}, which has 100,000 Colorectal cancer (CRC) patches from 86 patients. Each $224 \times 224$ pixel patch at $20\times$ magnification is annotated with one of nine distinct tissue class labels. 

For large-image augmentation, apart from TCGA-BRCA, we also use the BACH dataset~\cite{polonia_2020_3632035}. Introduced in the ICIAR 2018 Grand Challenge, the dataset contains $400$ H\&E-stained Breast Cancer images of size $2048 \times 1536$ pixels, evenly distributed across four categories: normal, benign, \textit{in-situ} carcinoma, and invasive carcinoma.

\subsection{Implementation details}
For all our experiments, we train the LDM on $256 \times 256$ pixel patches, following PathLDM~\cite{yellapragada2023pathldm}, which fine-tunes an ImageNet-trained~\cite{ILSVRC15} U-Net denoiser and uses a $4\times$ downsampling VQ-VAE, instead of the default LDM configuration. These modifications were deemed necessary for applying LDMs on large-image domains. 

We train our models on $6$ NVIDIA RTX 8000 GPUs, with a batch size of $100$ per GPU, utilizing code and pre-trained checkpoints from LDM~\cite{rombach2022high}.  We set the learning rate at $10^{-4}$ with a warmup of 10,000 steps. We apply DDIM with $50$ steps and a guidance scale of $1.75$ for both patch sampling and large-image generation. When generating large images we apply the patch diffusion model with a stride $>1$ depending on the desired target quality. 

\subsection{Image quality results}
\label{subsec:image_quality_metrics}
In Fig.~\ref{fig:real_synthetic}, we present synthetic patches and large images from our TCGA-BRCA and NAIP models, along with the corresponding references from which the self-supervised embeddings were extracted. We evaluate our method's per-patch and large-image generation quality by computing FID scores~\cite{heusel2017gans} using the Clean-FID implementation~\cite{parmar2021cleanfid}. We generate 10,000 patches ($256 \times 256$) and 3,000 large images ($1024 \times 1024$) from diffusion models trained on the TCGA-BRCA, TCGA-CRC, and NAIP datasets. Since our generative model requires a self-supervised conditioning vector $y$ (or multiple vectors $z$) for each synthetic image (or large image), we randomly sample embeddings from reference images in the training set to generate images for evaluation. 

For patches, we measure FID against the real images (``\textbf{Vanilla FID}"). For large images, we follow the evaluation strategy of MultiDiffusion~\cite{bar-tal2023multidiffusion} and use FID to compare the distribution of 256 $\times$ 256 crops from synthesized large images to that of real image patches of the same size (``\textbf{Crop FID}"). We also measure FID directly between the large images and ground truth data using CLIP~\cite{radford2021learning} (``\textbf{CLIP FID}").

To evaluate the similarity between the reference and generated large images, we resize from $1024 \times 1024$ to $256 \times 256$ and use an SSL model to extract embeddings. We compute the cosine similarity between the paired embeddings (``\textbf{Embedding Similarity}").

\textbf{Patch-level quality:} Our models achieve low patch FID scores across all datasets (Tab.~\ref{tab:fid}). For TCGA-BRCA patches, our ``Vanilla FID" of $\textbf{6.98}$ is comparable to the current state-of-the-art~\cite{yellapragada2023pathldm} ($7.64$ at $10\times$). We attain similar, low FID scores for the smaller CRC and NAIP datasets. 

\begin{table}[ht]
    {
\centering
\tabcolsep=0.1cm
\resizebox{1\linewidth}{!}{

\begin{tabular}{|c|c|c|ccc|}
\hline
\multirow{2}{*}{Dataset} & \multirow{2}{*}{\begin{tabular}[c]{@{}c@{}}\# Training\\ images\end{tabular}} & Patch level & \multicolumn{3}{c|}{Large image level}                                                                                                                                                                                 \\ \cline{3-6} 
                         &                                                                               & Vanilla FID & \multicolumn{1}{c|}{\begin{tabular}[c]{@{}c@{}}Crop \\ FID\end{tabular}} & \multicolumn{1}{c|}{\begin{tabular}[c]{@{}c@{}}CLIP  \\ FID\end{tabular}} & \begin{tabular}[c]{@{}c@{}}Emb \\ similarity\end{tabular} \\ \hline
BRCA 20x                 & 15M                                                                        & 6.98        & \multicolumn{1}{c|}{15.51}                                               & \multicolumn{1}{c|}{7.43}                                                 & 0.924                                                           \\ \hline
CRC 20x                  & 8M                                                                         & 6.78        & \multicolumn{1}{c|}{8.8}                                                 & \multicolumn{1}{c|}{7.34}                                                 & 0.938                                                           \\ \hline
NAIP                     & 667k                                                                         & 11.5        & \multicolumn{1}{c|}{43.76}                                               & \multicolumn{1}{c|}{6.86}                                                 & -                                                               \\ \hline
BRCA 5x                  & 976k                                                                         & 9.74        & \multicolumn{1}{c|}{-}                                                   & \multicolumn{1}{c|}{6.64}                                                    & -                                                               \\ \hline
\end{tabular}

}
}
    \caption{FID scores for our generated patch and large images. Our patch-level BRCA model is on par with SoTA~\cite{yellapragada2023pathldm} (7.64 at 10 $\times$). ``CLIP FID"  and ``Embedding Similarity" demonstrate our large images' realism and contextual accuracy.}
    \label{tab:fid}
\end{table}

\textbf{Synthetic large image quality:} We present the ``Crop FID", ``CLIP FID" and ``Embedding Similarity" results for large images in Table~\ref{tab:fid}. The ``Crop FID" of the large images is comparable to the ``Vanilla FID" on the BRCA and CRC datasets, showing that patches from synthesized large images have similar semantic content to the ground truth patches. We attribute the worse ``Crop FID" for the NAIP model to the limited number of samples available for both SSL and diffusion model training. The ``Crop FID" is consistently higher than the patch-level FID, which is expected as the large-image generation framework only approximates the distribution of the large images and does not have access to the true conditioning at every location. 

Our low ``CLIP FID" scores indicate that the generated large images are similar to real images when resized to $224 \times 224$ pixels. This indicates that our SSL-guided large-image generation successfully retains the larger-scale semantic arrangements of real data. For NAIP, our model is not as good at synthesizing high-frequency details, which explains the large discrepancy between ``Crop FID" and ``CLIP FID". Additionally, when comparing the ``CLIP FID" of large images synthesized by the $20\times$ BRCA model and resized to $5\times$, to images from a model trained directly on $5\times$ data, we see minimal difference (7.43 vs. 6.64).

We evaluate the contextual similarity between synthetic and reference large images for the BRCA and CRC data. We compute cosine similarity between large images using CTransPath embeddings. Our BRCA and CRC models demonstrate high ``Embedding Similarity" scores of $0.924$ and $0.938$, respectively, reflecting our framework's effectiveness in preserving the integrity of context and key features on the large-image scale.

\section{Image Augmentation Experiments}
\label{sec:data_aug}
As shown in Fig.~\ref{fig:real_synthetic}, apart from visual fidelity, the synthetic images preserve the intricate characteristics of the reference images, both on the patch and large-image scale. These characteristics include the nuanced textural elements, histological staining, and cell structure. This correspondence between real and synthetic images demonstrates the detailed and varied information captured by the self-supervised embeddings used as conditioning. In conjunction with the high ``Embedding Similarity", it justifies using variations of images generated from SSL embeddings for patch and large-image level data augmentation.

Having a powerful generative model enables us to perform data augmentation for tasks where we can control the augmented image label using conditioning. In our setting, our diffusion models are not trained with class labels; instead, we synthesize a novel image using the conditioning from a reference patch or region.

We assume that i) the self-supervised embedding used as conditioning contains information about the target label and ii) the diffusion-generated variations of an image do not alter this target label information. We validate both assumptions experimentally, by augmenting training sets on patch and large-image level tissue classification tasks, including a Multiple Instance Learning (MIL) task.

\begin{table*}[ht]
    {
\centering
\resizebox{1\linewidth}{!}{
\begin{tabular}{|c|ccccc|ccccc|}
\hline
        & \multicolumn{5}{c|}{TCGA-BRCA Subtyping}                                                                                                                                                                                                                                                                                                                                                           & \multicolumn{5}{c|}{TCGA-BRCA HRD}                                                                                                                                                                                                                                                                                                                                                                 \\ \hline
Method  & \multicolumn{1}{c|}{\begin{tabular}[c]{@{}c@{}}1\%\\ Real\end{tabular}} & \multicolumn{1}{c|}{\begin{tabular}[c]{@{}c@{}}1\% Real\\ + synthetic\end{tabular}} & \multicolumn{1}{c|}{\begin{tabular}[c]{@{}c@{}}10\%\\  Real\end{tabular}} & \multicolumn{1}{c|}{\begin{tabular}[c]{@{}c@{}}10\% Real \\ + synthetic\end{tabular}} & \begin{tabular}[c]{@{}c@{}}20\%\\  Real\end{tabular} & \multicolumn{1}{c|}{\begin{tabular}[c]{@{}c@{}}1\%\\ Real\end{tabular}} & \multicolumn{1}{c|}{\begin{tabular}[c]{@{}c@{}}1\% Real\\ + synthetic\end{tabular}} & \multicolumn{1}{c|}{\begin{tabular}[c]{@{}c@{}}10\% \\ Real\end{tabular}} & \multicolumn{1}{c|}{\begin{tabular}[c]{@{}c@{}}10\% Real \\ + synthetic\end{tabular}} & \begin{tabular}[c]{@{}c@{}}20\% \\ Real\end{tabular} \\ \hline
CLAM-SB & \multicolumn{1}{c|}{0.725}                                              & \multicolumn{1}{c|}{\textbf{0.812}}                                                 & \multicolumn{1}{c|}{0.886}                                                & \multicolumn{1}{c|}{\textbf{0.898}}                                                   & 0.91                                                 & \multicolumn{1}{c|}{0.603}                                              & \multicolumn{1}{c|}{\textbf{0.644}}                                                 & \multicolumn{1}{c|}{0.649}                                                & \multicolumn{1}{c|}{\textbf{0.765}}                                                   & 0.787                                                \\ \hline
DSMIL   & \multicolumn{1}{c|}{0.609}                                              & \multicolumn{1}{c|}{\textbf{0.659}}                                                 & \multicolumn{1}{c|}{0.838}                                                & \multicolumn{1}{c|}{\textbf{0.856}}                                                   & 0.905                                                & \multicolumn{1}{c|}{0.517}                                              & \multicolumn{1}{c|}{\textbf{0.554}}                                                 & \multicolumn{1}{c|}{0.563}                                                & \multicolumn{1}{c|}{\textbf{0.639}}                                                   & 0.669                                                \\ \hline
\end{tabular}}
}
    \caption{The inclusion of synthetic data consistently enhances AUC across various MIL architectures and BRCA tasks. The dataset contains 1000 real images, so ``10\% + synthetic" indicates training with 100 real and 100 synthetic WSIs, with the remainder used for testing.}
    \label{tab:mil_augmentation}
\end{table*}

\textbf{Large-image augmentation on TCGA-BRCA:} We examine two histopathology slide-level binary classification tasks on TCGA-BRCA: BRCA Subtyping (Invasive Ductal Carcinoma (IDC) vs Invasive Lobular Carcinoma (ILC)) and HRD prediction. We utilize a minimal dataset, just $10\%$ of real WSI data ($100$ WSIs), to train MIL algorithms. We generate an equal set of synthetic images for $100$ additional WSIs using training set images as reference.

We employ 10-fold cross-validation to divide our dataset into training and testing segments. Within each fold, two multiple instance learning (MIL) models, CLAM-SB~\cite{lu2021data} and DSMIL~\cite{li2021dual}, are trained on two sets: one with real data and another combining real and synthetic data. To train the MIL models, we extract features using the CTransPath ViT~\cite{wang2021transpath}. The results, detailed in Table~\ref{tab:mil_augmentation}, indicate that models trained on the augmented datasets consistently surpass their real-only counterparts, regardless of the MIL algorithm used. This demonstrates the value of the synthetic images generated by our method, confirming their efficacy as comparable to real images for training purposes.

\textbf{Large-image augmentation on BACH:} We double the training set of BACH~\cite{polonia_2020_3632035} by adding as many synthetic large-images, produced by the TCGA-BRCA diffusion model. From each $2048 \times 1536$ pixel training set image, we extract a $8 \times 6$ SSL embedding grid to generate a variation with the same label. For classification, we employ a $\text{ConvNeXt V2}\_\text{huge}$~\cite{woo2023convnext} model pre-trained on ImageNet. We train a 2-layer MLP classifier on top of the penultimate layer features and evaluate it on the official test set.

The results, presented in Table~\ref{tab:bach}, reveal a notable improvement in classifier performance, from $78\%$ to $83\%$. This improvement again confirms the high quality of our synthetic data while also highlighting the versatility of our apporach. Despite being trained on $256 \times 256$ patches from TCGA-BRCA, the model generalizes to produce realistic large images from a completely different dataset. We attribute this generalization capability to the expressiveness of the SSL features and the potency of the diffusion model to accurately portray them in images. While our model does not reach the current SoTA accuracy of $87\%$, achieved by an ensemble approach~\cite{chennamsetty2018classification}, the simplicity and ability to integrate with other models make for a noteworthy contribution.

\begin{table}[ht]
    {
\centering
\begin{subtable}[t]{0.48\columnwidth}{
\centering
\resizebox{1\linewidth}{!}{
    \begin{tabular}{|c|c|}
    \hline
    Training Data   & Test Acc          \\ 
    \hline
    Real             & 78 \%              \\ 
    
    Synthetic        & 70 \%             \\ 
    
    Real + synthetic & 83 \%   \\ 
    \hline
     SoTA  \cite{chennamsetty2018classification}  & \textbf{87 \%}    \\ 
    \hline
    \end{tabular}}
\caption{}
    \label{tab:bach}
}
\vspace{-2mm}
\end{subtable}
\hfill
\begin{subtable}[t]{0.48\columnwidth}{
    \centering
\resizebox{1\linewidth}{!}{
    \begin{tabular}{|c|c|}
    \hline
    Training Data       & Val Acc         \\ \hline
    Real             & 93.8 \%            \\ 
    Synthetic        & 90.19 \%          \\ 
    Real + synthetic & \textbf{96.27 \%}  \\ \hline
    SoTA \cite{KUMAR2023104172} & 96.26 \% \\ \hline
    \end{tabular}}
    \caption{}
    \label{tab:nct-crc}
}
\vspace{-2mm}
\end{subtable}
}
    \caption{Our data augmentations provide notable improvements for the BACH (a) and CRC-VAL-HE (b) datasets. Notably, the diffusion training data \emph{does not overlap} with the data of the augmented datasets.}
    \label{tab:image_aug}
\end{table}

\textbf{Patch-level image augmentation:} We further investigate our out-of-distribution generalization capabilities by augmenting the NCT-CRC dataset~\cite{kather_2018_1214456}. We leverage our diffusion model trained on the TCGA-CRC data, which does not overlap with NCT-CRC, conditioned on embeddings from an iBOT ViT~\cite{filiot2023scaling}. We generate an augmented dataset of equal size to the original by using the SSL embedding of every patch in the NCT-CRC training set to synthesize a corresponding image of the same label.

We train an ImageNet pre-trained ResNet-50~\cite{he2016deep} network on three splits: real images only, synthetic images only, and a combination of both. Evaluation on the CRC-VAL-HE-7K test set demonstrates a significant performance increase when synthetic data is introduced, with classifier accuracy rising from \textbf{93.8\%} to \textbf{96.27\%} as presented in Table~\ref{tab:nct-crc}. We match the current SoTA~\cite{KUMAR2023104172}, which used an ensemble of deep models, with a model agnostic approach. As in the previous experiment, the diffusion model, now trained on the significantly smaller TCGA-CRC data, can also effectively synthesize images from a completely different dataset by only controlling the self-supervised conditioning. 

\section{Text-to-large image synthesis}
\label{sec:other_exp}
\label{subsec:text_cond_gen}
For previous tasks, our large image generation approach synthesizes variations of an existing set of images from the pre-computed self-supervised embeddings. In Sec.~\ref{sec:controllable_synth} we discussed how to control large image generation with auxiliary signals from any domain (class labels, text captions, etc.), by training models $p (z \mid c)$ that can be combined with the embedding-conditioned image generation. We demonstrate controllable image synthesis with text-to-large image generation experiments on histopathology and satellite data. We measure the similarity between synthetic images and the text prompts used in generating them using vision-language models (VLMs).

\textbf{Text-to-large histopathology images:} We utilize the CRC and BRCA diffusion models to generate $1024 \times 1024$ pixel images from text prompts. We first construct training sets by pairing $4 \times 4$ SSL embedding grids $z$ from large BRCA and CRC images, with their corresponding Quilt~\cite{ikezogwo2023quilt} image embeddings $c$. We then train an auxiliary diffusion model to sample $p (z \mid c)$. During inference, we use a text embedding $c'$ as a proxy for the image embedding, to sample $z$ and synthesize a large image. To bridge the gap between the image and text Quilt embeddings~\cite{liang2022mind, nukrai2022text} we perturb the image embeddings when training the auxiliary diffusion model with Gaussian noise of variance $\sigma^2=0.1$.

\begin{figure}[ht]
    \centering
    \includegraphics[width=\linewidth]{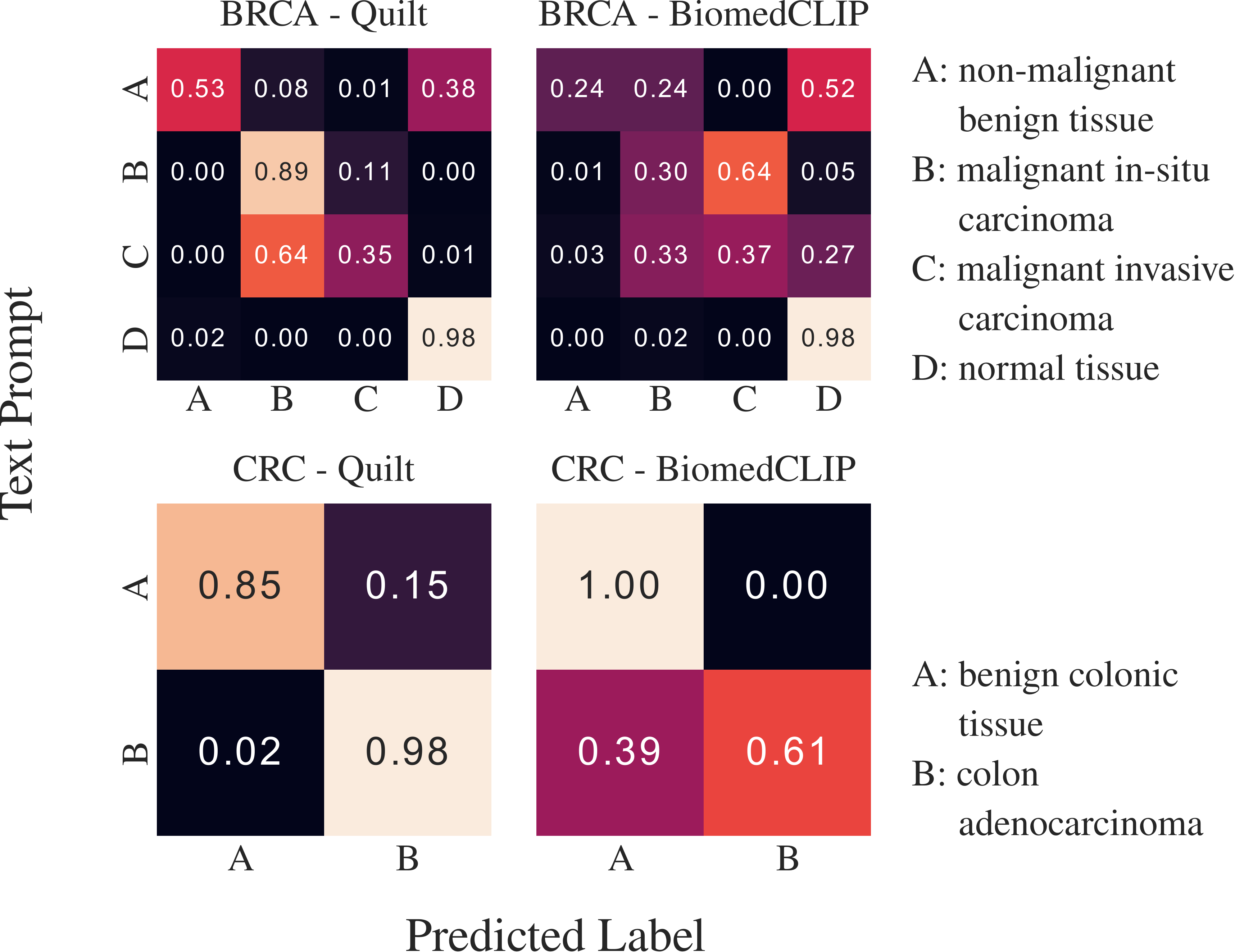}
    \caption{Confusion matrix of zero-shot classification for novel TCGA-CRC and TCGA-BRCA synthetic images.}
    \label{fig:novel_confusion}
\end{figure}

To evaluate the text-to-large image pipeline, we generate images from a pre-defined set of classes described in natural language; non-malignant benign tissue, malignant in-situ carcinoma, malignant invasive carcinoma, normal breast tissue for BRCA and colon adenocarcinoma, benign colonic tissue for CRC. We construct the confusion matrix of zero-shot classifiers on the synthesized data. We used two different VLMs as zero-shot classifiers, Quilt and BiomedCLIP~\cite{zhang2023large}. The results presented in Fig.~\ref{fig:novel_confusion} demonstrate our ability to synthesize images consistent with the text prompts. The capabilities of the VLM used, limit our synthetic image generation. The lower performance of BRCA vs. CRC is consistent with the results reported in Quilt~\cite{ikezogwo2023quilt}. Furthermore, we asked an expert pathologist to classify 100 synthetic CRC images as benign or adenocarcinoma images. Their evaluation showed an \textbf{89.9 \%} agreement rate with the labels used for image generation, indicating consistency in our text-to-large image pipeline. We show examples of synthesized images in the supplementary.

\textbf{Text-to-large satellite images:} To synthesize novel satellite images we first create a training set of 30k $1024 \times 1024$ pixel large NAIP images, and pair them with captions from a BLIP model~\cite{li2022blip}. We train a diffusion model to sample the $4 \times 4$ SSL embeddings $z$ from the captions $c$. To evaluate, we create a separate set of 1000 NAIP image-caption pairs and measure the CLIP similarity between generated images and the given captions. We achieve a CLIP similarity score of \textbf{0.22}, showing that we can effectively learn the mapping from text to large images with this hierarchical approach. Although our CLIP similarity is slightly worse than the scores reported for text-to-image Stable Diffusion models ($>0.24$)~\cite{podell2023sdxl}, we expect this drop in performance as we trained with machine-generated captions. Training and generated image-caption pairs are provided in supplementary.

\section{Combining self-supervised embeddings with diffusion}
\label{sec:expt_feat_mil}
We further evaluate our patch-generating models by posing the question \textit{does the diffusion model learn more about the data than the self-supervised learning model?} We hypothesize that by combining the pre-trained self-supervised embedding with the denoising task we can improve the learned representations of the data, leading to better performance in downstream tasks, which are performed with features from self-supervised learning. To validate this hypothesis, we utilize the trained diffusion model as a feature extractor and apply a Multiple-instance learning (MIL) approach for the slide-level classification task of subtyping Breast Cancer. For each patch in a whole-slide image (WSI) we first extract the self-supervised embeddings, which are then used as conditioning to obtain features from the denoiser's U-Net bottleneck layer at a fixed timestep $t=50$. 

In Tab.~\ref{tab:mil_diff_feat}, we evaluate the effectiveness of our novel fusion of generative diffusion and self-supervised embeddings. We compare the performance of MIL algorithms~\cite{lu2021data, li2021dual} using features derived from our approach at $20\times$ and $5\times$ magnification ($\text{LDM}_{\text{HIPT}/\text{CTransPath}}$) against using only the self-supervised conditioning (HIPT/ CTransPath). We used a 10-fold cross-validation strategy consistent with the data splits from HIPT~\cite{chen2022scaling}, training the MIL algorithms on both the full dataset (100\%) and a reduced subset (25\%). 

The results indicate that integrating self-supervised features into the diffusion model as conditioning leads to learning better representations and improves whole-slide classification. By fusing the generative knowledge from the diffusion process with the discriminative capabilities of the self-supervised embeddings, we construct a successful model for both discriminative and generative tasks.

\begin{table}[ht]
    {
    \centering
    \tabcolsep=0.11cm
    \resizebox{1\linewidth}{!}{
    \begin{tabular}{|c|c|cc|cc|}
    \hline
    &                                        & \multicolumn{2}{c|}{25\% training} & \multicolumn{2}{c|}{100\% training}  \\
    Mag & Features                 & CLAM-SB          & DSMIL           & CLAM-SB           & DSMIL            \\ 
    \hline
    \multirow{2}{*}{$20\times$} & HIPT        & 0.788            & 0.784           & 0.861             & 0.839            \\
    & $\text{LDM}_\text{HIPT}$               & \textbf{0.842}   & \textbf{0.795}  & \textbf{0.908}    & \textbf{0.894}   \\ 
    \hline
    \multirow{2}{*}{$5\times$} & CTransPath & 0.900            & 0.896           & 0.919             & 0.910            \\
    & $\text{LDM}_\text{CTransPath}$         & \textbf{0.913}   & \textbf{0.905}  & \textbf{0.923}    & \textbf{0.936 }  \\ 
    \hline
    \end{tabular}
    }
}
    \caption{10-fold cross-validation AUC for BRCA Histological subtyping. $\text{LDM}_\text{HIPT}$ denotes diffusion features conditioned on HIPT embeddings. The fusion of SSL and diffusion features outperforms the SSL features by themselves. }
    \label{tab:mil_diff_feat}
\end{table}

\section{Conclusion}
\label{sec:conc}
We presented a novel approach to training diffusion models in large-image domains, such as digital histopathology and remote sensing. We overcome the need for fine-grained annotation by introducing self-supervised representation guided diffusion models, achieving remarkable image synthesis results on the patch level. Our approach also enables us to synthesize high-quality large images, where we have the ability to dictate the global structure by controlling the spatial arrangement of the conditions. We evaluated the usefulness of our synthetic images on a number of patch and large image-level tasks, as well as introduced a text-to-large image generation framework. Naively augmenting whole-slide images is a time-consuming process. We leave to future work the exploration of adaptive augmentation strategies that choose which image parts to augment. We believe these results illustrate the great potential for this technology to lead to bespoke foundational models for specialized domains, comparable to existing models for natural images.

\paragraph{Acknowledgements} This research was partially supported by NCI awards 5U24CA215109, 1R21CA258493-01A1, UH3CA225021, NSF grants IIS-2123920, IIS-2212046 and Stony Brook Profund 2022 seed funding. We thank Rajarsi Gupta for his valuable feedback.

{
    \small
    \bibliographystyle{ieeenat_fullname}
    \bibliography{main}
}

\maketitlesupplementarysingle

\section{Annotation costs}
\begin{figure}[ht]
    \centering
    \includegraphics[width=\textwidth]{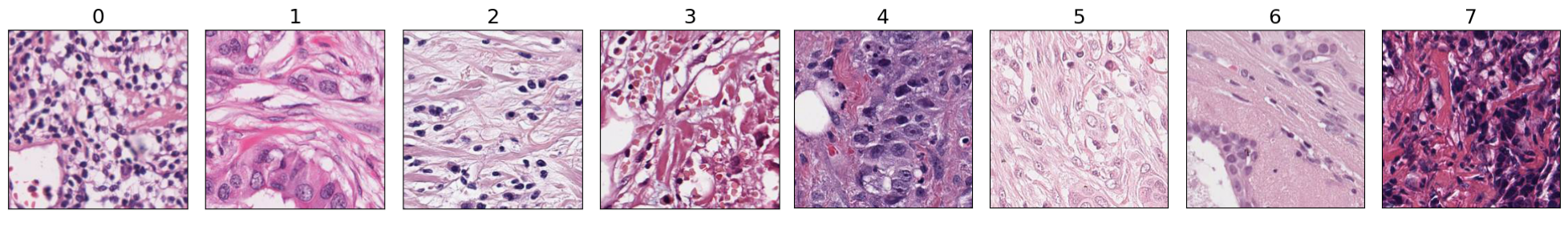}
    \caption{Examples of patches annotated by an expert pathologist. For each image, the pathologist required 5-10s to provide a brief, detailed description of the features visible. Annotating the entirety of TCGA in this manner is a colossal task.}
    \label{fig:pathologist_annot}
\end{figure}

We asked a pathologist to annotate patches from TCGA-BRCA to estimate the cost of detailed per-patch annotation for the entirety of the dataset. We presented the $20\times$ magnification patches of Fig.~\ref{fig:pathologist_annot} and requested them to \emph{``write a brief description for each of the following patches"}. An expert pathologist required approximately 5-10 seconds to identify features and describe the patches. Therefore, for the entire 15M patches of TCGA-BRCA, it would take $\approx40000$ hours to provide full per-patch annotations. This training dataset is small compared to the volume of data used in large studies, e.g. 10k whole slide images or approximately $10\times$ the number of TCGA-BRCA data. Employing expert pathologists to annotate these vast amounts of data is prohibitively expensive and, therefore, practically infeasible at the scale at which we want to apply these models.

\section{Out-of-distribution augmentation examples}
\begin{figure}[ht]
    \centering
    \includegraphics[width=\textwidth]{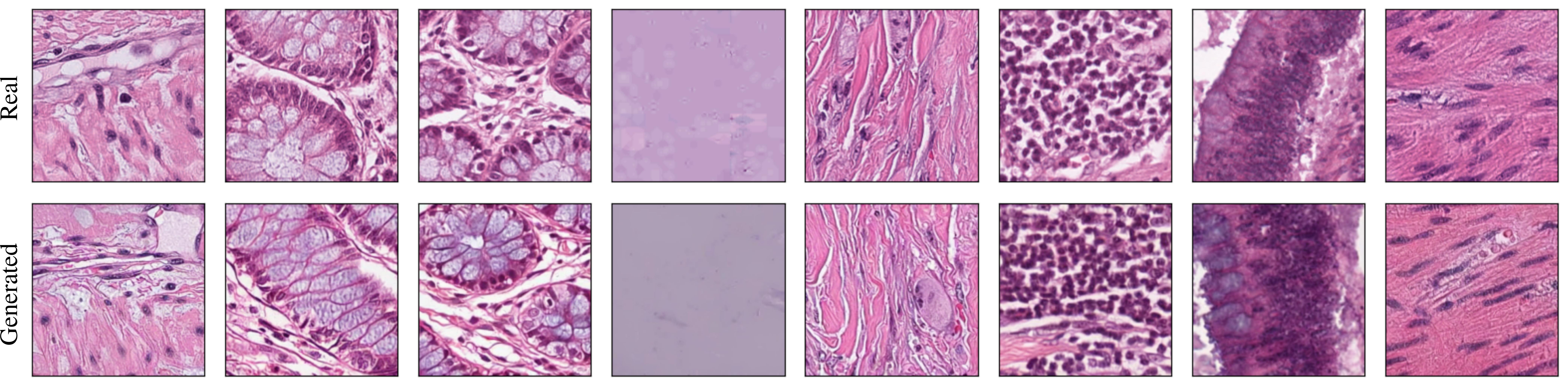}
    \caption{Synthetic images from NCT-CRC. For each generated image we extract the SSL embedding from a real reference image, taken from NCT-CRC-HE-100K, and generate a patch using the TCGA-CRC model. The synthesized patches are similar to the reference in both appearance and semantics. The TCGA-CRC model was never trained on data from the NCT-CRC-HE-100K dataset.}
    \label{fig:nct_crc_examples}
\end{figure}

\begin{figure}[ht]
    \centering
    \includegraphics[width=\textwidth]{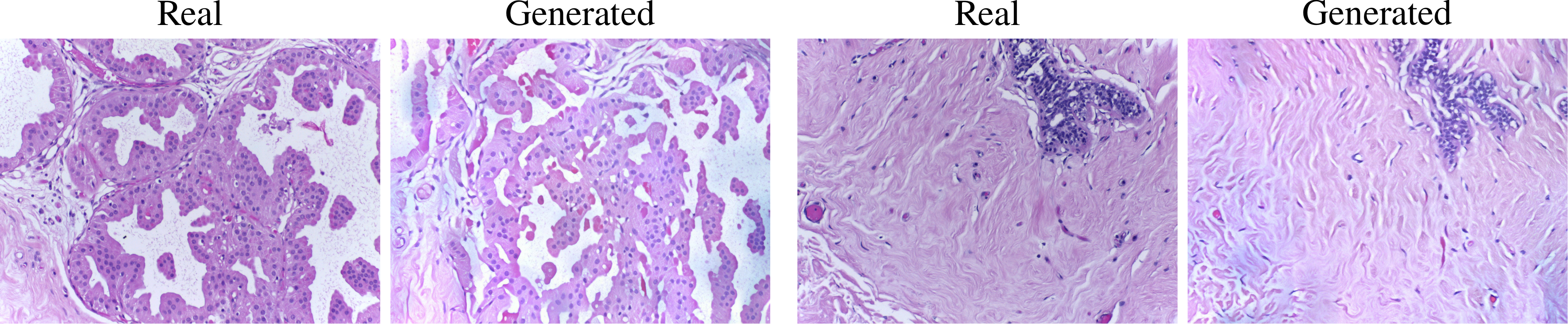}
    \caption{Examples of generated images from BACH. For each generated image we extract the SSL embeddings from a reference image, taken from the BACH dataset. We generate the large image using the TCGA-BRCA model. Although the appearance between the reference and generated images is slightly different, the large images maintain the global semantics. The TCGA-BRCA model was never trained on data from the BACH dataset.}
    \label{fig:bach_examples}
\end{figure}

In Fig.~\ref{fig:nct_crc_examples} and Fig.~\ref{fig:bach_examples} we show out-of-distribution examples of generated images from the NCT-CRC and BACH datasets, along with the reference image from which the SSL embeddings were extracted. For NCT-CRC, it is evident that the synthetic patches follow the semantics and appearance of the real patches used. Regarding BACH, we find the appearance to be slightly different between the real large images and our synthetic large images, but we see that the semantic contents are mostly left unchanged. This is also validated by our augmentation experiments in the main text, where we improve the classification accuracy with synthetic BACH data. In both cases, our SSL-conditioned diffusion models exhibit impressive generalization capabilities by only modifying the conditioning provided to them. Given that generalization is an essential property for building foundation models, we believe that our work is an important step towards this direction for large image domains such as digital histopathology and remote sensing.

\section{NCT-CRC augmentation additional results}
We expand the results of Table~\ref{tab:nct-crc} in the main text by evaluating the classification accuracy on the CRC-VAL-HE-7K test set with more synthetic data. As shown in Table~\ref{tab:nct_crc_aug_more}, expanding the dataset with more than $2\times$ synthetic data does not improve the performance further. Adding more synthetic data ends up hurting the classifier, which we attribute to the dilution of the real data with the imperfect, synthetic variations that we generate with our diffusion model. Even so, the final classification accuracy with $5\times$ synthetic data is still higher than the baseline that only uses real images.

\begin{table}[ht]
    \centering
    \begin{tabular}{|c|c|}
        \hline
        Training Data              & Val Acc \\
        \hline
        Real                       & 93.8\%  \\
        \hline
        Real + $1\times$ Synthetic & 96.27\% \\
        Real + $2\times$ Synthetic & \textbf{96.55\%} \\
        Real + $3\times$ Synthetic & 95.52\% \\
        Real + $5\times$ Synthetic & 95.59\% \\
        \hline
    \end{tabular}
    \caption{Classification accuracy on the CRC-VAL-HE-7K test set for different quantities of synthetic data. Expanding the training data with more than $2\times$ synthetic samples does not improve the classification accuracy further.}
    \label{tab:nct_crc_aug_more}
\end{table}

\section{Memory requirements}
We propose using an LDM trained on $256 \times 256$ pixel patches to generate large images of size $1024 \times 1024$. In Table \ref{tab:memory}, we compare the requirements of training an LDM directly on $1024 \times 1024$ pixel large images, instead of our patch-based approach. We use the same $4\times$ downsampling factor for the first stage VAE and employ a single RTX 6000 GPU for benchmarking purposes. Training a diffusion model on $1024 \times 1024$ resolution TCGA-BRCA images requires an order of magnitude more time than our patch-based approach for the same number of iterations. However, with a reduced batch size, we also empirically know that it would require more training iterations for the model to converge. We argue that since our approach can be used to generate large images without significant loss in quality, our patch-based model is the more efficient solution.

\begin{table}[ht]
    \centering
    \begin{tabular}{|c|c|c|}
    \hline
    Training Method                 & Maximum batch size & Training time  per epoch \\ \hline Ours (256 $\times$ 256 patches)   & 100                                                           & 45 hr                                                                \\ \hline LDM on  1024 $\times$ 1024 images & 4                                                             & 300 hr                                                           \\ \hline
    \end{tabular}
    \caption{Training a diffusion model on large images is computationally expensive and takes an order of magnitude more time.}
    \label{tab:memory}
\end{table}

\begin{table}[ht]
    {
    \centering
    \tabcolsep=0.1cm
    \begin{minipage}[t]{0.45\linewidth}
    \centering
    \begin{tabular}{|l|c|}
        \hline
        Conditioning                    & Patch FID               \\
        \hline
        None                            & 25.62                   \\  
        ImageNet ViT-B/16               & 13.29                   \\
        CLIP \cite{radford2021learning}                     & 16.07                   \\
        HIPT \cite{chen2022scaling}                      & \textbf{6.98}           \\
        \hline
    \end{tabular}
    \caption{FIDs when using different representations as conditions.}
    \label{tab:ssl_ablation}
    \end{minipage}
    \hfill
    \begin{minipage}[t]{0.45\linewidth}
    \centering
    \begin{tabular}{|c|c|c|c|}
        \hline
        \multirow{2}{*}{Stride} & Time/ & Crop & CLIP \\
                                & Image & FID  & FID  \\
        \hline
        4                       & 15m   & \textbf{12.66} &  \textbf{7.31} \\
        8                       & 4m    & 14.69          &  7.37          \\
        16\textsuperscript{*}   & 1m    & 15.51          &  7.43          \\
        32                      & 20s   & 15.60          &  8.09          \\
        \hline
    \end{tabular}
    \caption{Large image generation parameters ablation. By \textsuperscript{*} we denote the stride used in the main text experiments.}
    \label{tab:large_image_ablation}
    \end{minipage}
    }
\end{table}

\section{Using different SSL encoders}

We extend the TCGA-BRCA $20\times$ model of Table~\ref{tab:fid} with additional patch-level FID values, obtained by using different embeddings as conditioning (Table \ref{tab:ssl_ablation}). The pathology-specific HIPT performs best, suggesting that the domain expressivity of the embedding used as conditioning affects image generation quality. We conjecture that worse patch quality also hurts large image metrics.

\section{Large image generation details}
To generate large images we use DDIM \cite{song2020denoising} with 50 inference steps and a classifier-free guidance weight of 3.0. The SSL conditioning (384 or 768 dimensional vector depending on the SSL model) is first normalized with the $L_2$ norm and then projected to a 512-dim vector using a linear layer. The null token for the classifier-free guidance is represented by replacing the SSL embedding with a vector of all 0s. The conditioning is applied to the U-Net model using cross-attention, similar to other LDM conditioning mechanisms \cite{rombach2022high}.

The LDM is applied to patches in the large image with a stride of 16. Using a larger stride leads to tiling artifacts, whereas a smaller stride increases the computational cost without much difference in the synthesized image quality. In Table \ref{tab:large_image_ablation} we provide an ablation study of the large image generation parameters. We synthesize $1024\times1024$ px images from TCGA-BRCA with different strides, using 50 steps of diffusion, on an NVIDIA RTX 6000, showing that larger strides require fewer forward passes (less time) but produce worse results. 

For each location $i,j$ at which we want to apply the diffusion model, we interpolate the 4-nearest embeddings to get the conditioning $\lambda_{i,j}$. We found that spherical linear interpolation (slerp), weighted by the distance of $i,j$ to the centers of its four neighbors, worked best for interpolating the high-dimensional, normalized SSL embeddings.

When averaging the diffusion updates we first applied a Gaussian kernel to downweight the pixels at the edges of the patch. This helps with unwanted tiling artifacts as we 'trust' the diffusion updates in the center more than the edges of a patch. Likewise, when decoding the large image latents into images, we used a stride of 16 with the Gaussian kernel weighting, to eliminate tiling artifacts in the decoded images.

For the text-to-large image experiments, we trained an auxiliary diffusion model to sample a $4 \times 4$ grid of embeddings given the text conditioning. We used a small convolutional network with residual layers to implement the diffusion model. The timestep conditioning was concatenated to the input grid of embeddings. The network directly predicted the final embeddings from the conditioning and current noisy embedding grid, instead of predicting the noise added. For TCGA-BRCA and TCGA-CRC, the text conditioning is a single 512-dim Quilt embedding vector. For NAIP, we used a frozen CLIP \cite{radford2021learning} text encoder to extract features from the text captions and used them as conditioning. For the diffusion process, we used 1000 steps with a linear schedule, as in \cite{ho2020denoising}. Additionally, when sampling embeddings from text for TCGA-BRCA and TCGA-CRC we used negative prompting \cite{liu2022compositional} to further separate the different types of images during generation.

\section{Text-to-large image generation examples}
\begin{figure}[ht]
    \centering
    \includegraphics[width=\textwidth]{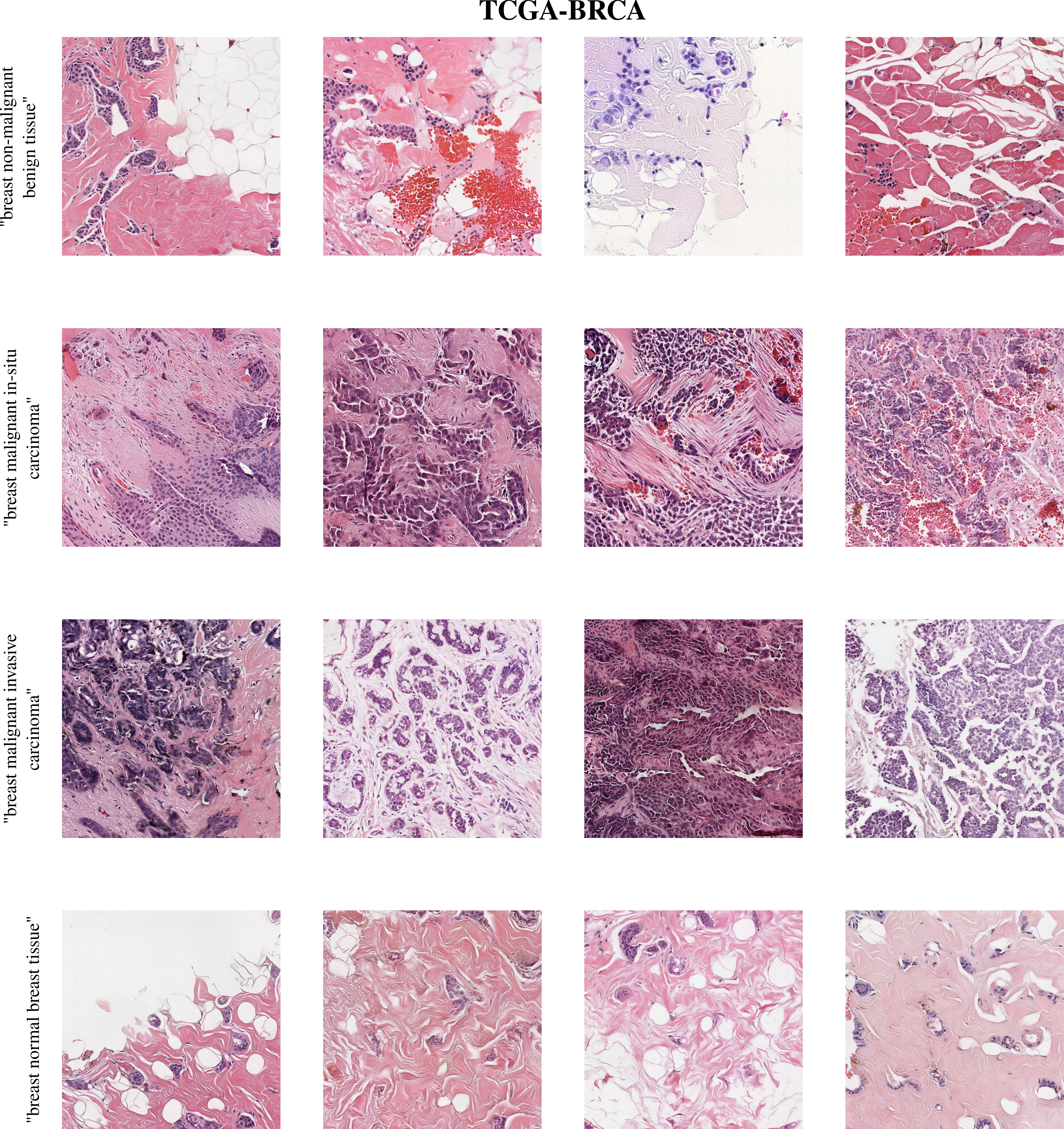}
    \caption{Generated samples from TCGA-BRCA along with the text prompt used. We use the zero-shot classification prompts from Quilt \cite{ikezogwo2023quilt} to generate the embeddings. Images are at $1024 \times 1024$ pixels resolution.}
    \label{fig:brca_examples}
\end{figure}

\begin{figure}[ht]
    \centering
    \includegraphics[width=\textwidth]{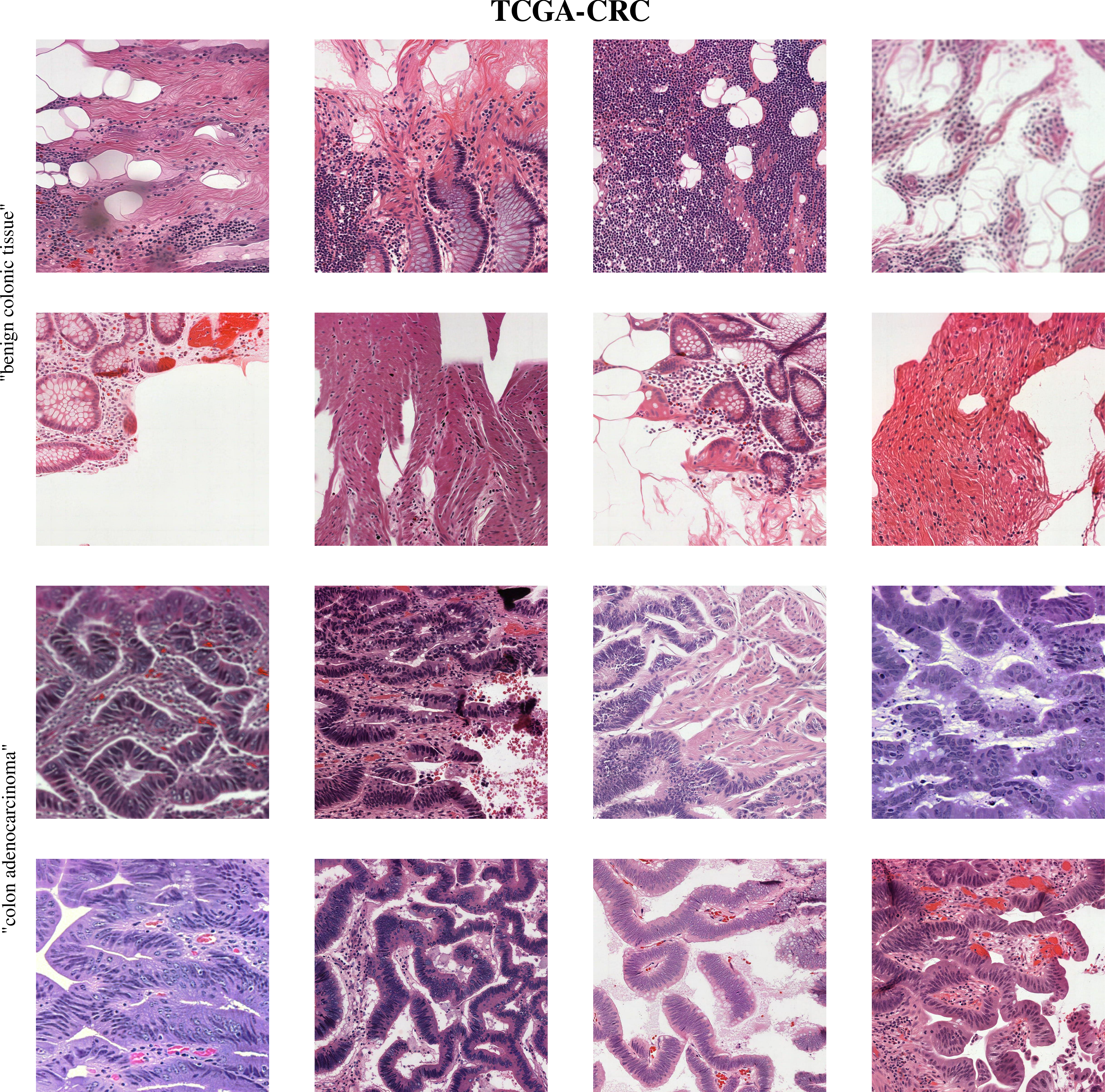}
    \caption{Generated samples from TCGA-CRC along with the text prompt. We use the zero-shot classification prompts from Quilt \cite{ikezogwo2023quilt} to generate the embeddings. Images are at $1024 \times 1024$ pixels resolution.}
    \label{fig:crc_examples}
\end{figure}

In Fig.~\ref{fig:brca_examples} we present generated images from the TCGA-BRCA model and the text prompts used in generating them. We borrow the text prompts from the zero-shot classification experiments of \cite{ikezogwo2023quilt}. As discussed for the confusion matrix (Fig.~\ref{fig:novel_confusion}), the vision-language model's capabilities limit the quality of our results. The model seems to be able to only differentiate between \textit{non-malignant} / \textit{normal} and \textit{malignant}, which is expected since the zero-shot classification accuracy of Quilt on breast cancer images is around 40\%. In contrast, for the CRC data where accuracy is around 90\%, our text-to-large image generation performs better. In Fig.~\ref{fig:crc_examples} we present such synthetic samples from TCGA-CRC.

\begin{figure}[ht]
    \centering
    \includegraphics[width=0.95\textwidth]{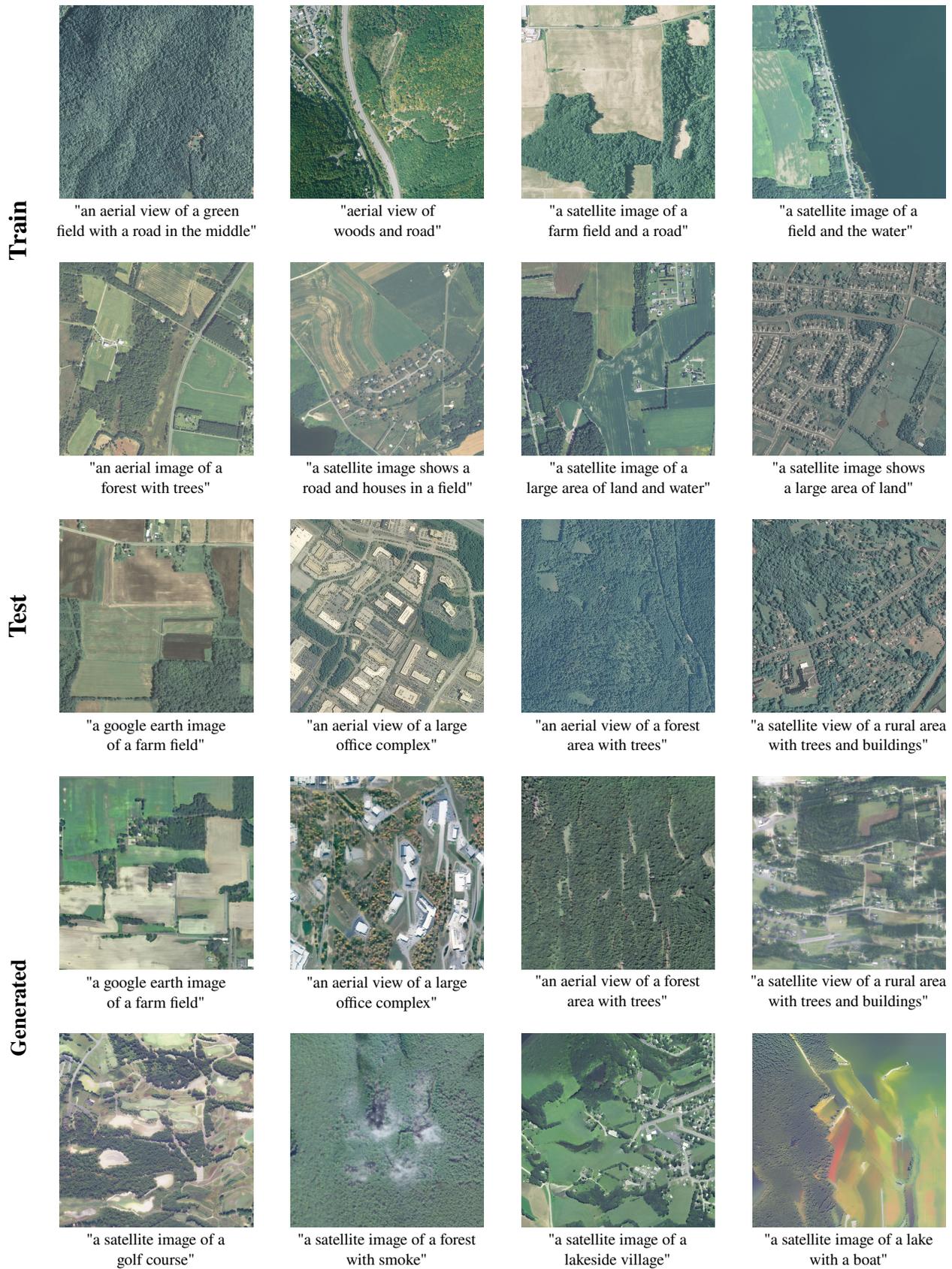}
    \caption{Examples of training, test and generated text-to-large satellite images. Images are at $1024 \times 1024$ pixels resolution.}
    \label{fig:sat_examples}
\end{figure}

To train the satellite text-to-large image auxiliary diffusion model we generated a synthetic set of image-caption pairs using BLIP \cite{li2022blip}. For training, we created a set of 30k large images ($1024 \times 1024$ pixels) with 4 captions for each, whereas for the test set, we used a single caption for evaluation. In Fig.~\ref{fig:sat_examples} we present images from the training and test sets as well as generated samples along with their text prompts. We see that although the training captions are far from perfect, we are able to generate test set images consistent with the prompts used. Even though our training set is tiny, we see interesting generalization capabilities when using 'unusual' prompts, such as \textit{"a satellite image of a forest
with smoke"}, where the model tries to add clouds to mimic the "smoke" seen from a satellite image. This generalization can be attributed to both the expressivity of the SSL embeddings used in synthesizing the images and the usage of a pre-trained CLIP text encoder to interpret the captions.

\section{Pathologist evaluation}
\begin{figure}[ht]
    \centering
    \includegraphics[width=0.5\textwidth]{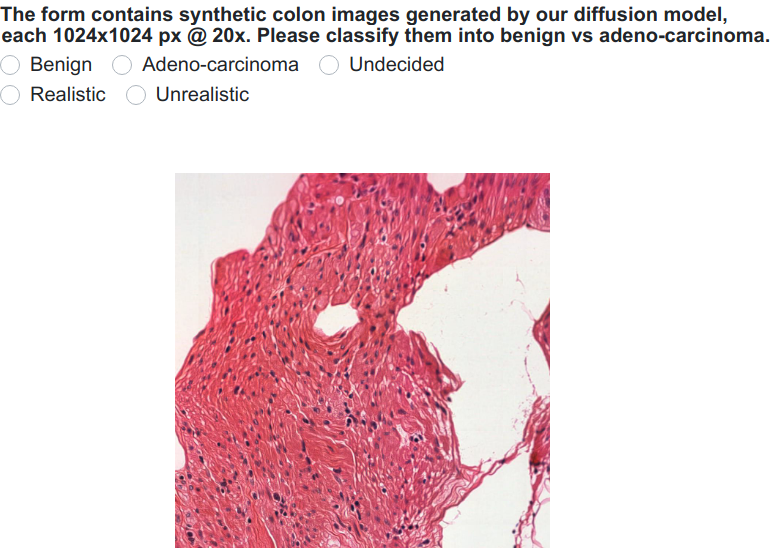}
    \caption{Pathologist evaluation UI. We presented synthetic images to an expert pathologist and asked them to evaluate them. The results showed 89.9\% agreement between the text prompts used to generate the images and the pathologist's assessment.}
    \label{fig:pathologist_eval}
\end{figure}

We designed a simple user interface where we presented large TCGA-CRC images generated from text prompts and asked an expert pathologist to evaluate them (Fig.~\ref{fig:pathologist_eval}). The model generated an image using one of two text prompts: \textit{"benign colonic tissue"} or \textit{"colon adenocarcinoma"}. We asked a pathologist to evaluate by categorizing the images as \textit{benign} / \textit{adenocarcinoma} / \textit{undecided} as well as assigning a \textit{realistic} / \textit{unrealistic} label. For a total of 100 images, the final agreement between text prompts and pathologist labels was 89.9\%, with 61\% of the images marked as realistic. This clearly illustrates the applicability of our proposed method; an auxiliary diffusion model that generates the SSL conditioning from any related modality can be chained with our patch-based diffusion to synthesize coherent large images.

\section{Embedding resolution}
\begin{figure}[ht]
    \centering
    \includegraphics[width=\textwidth]{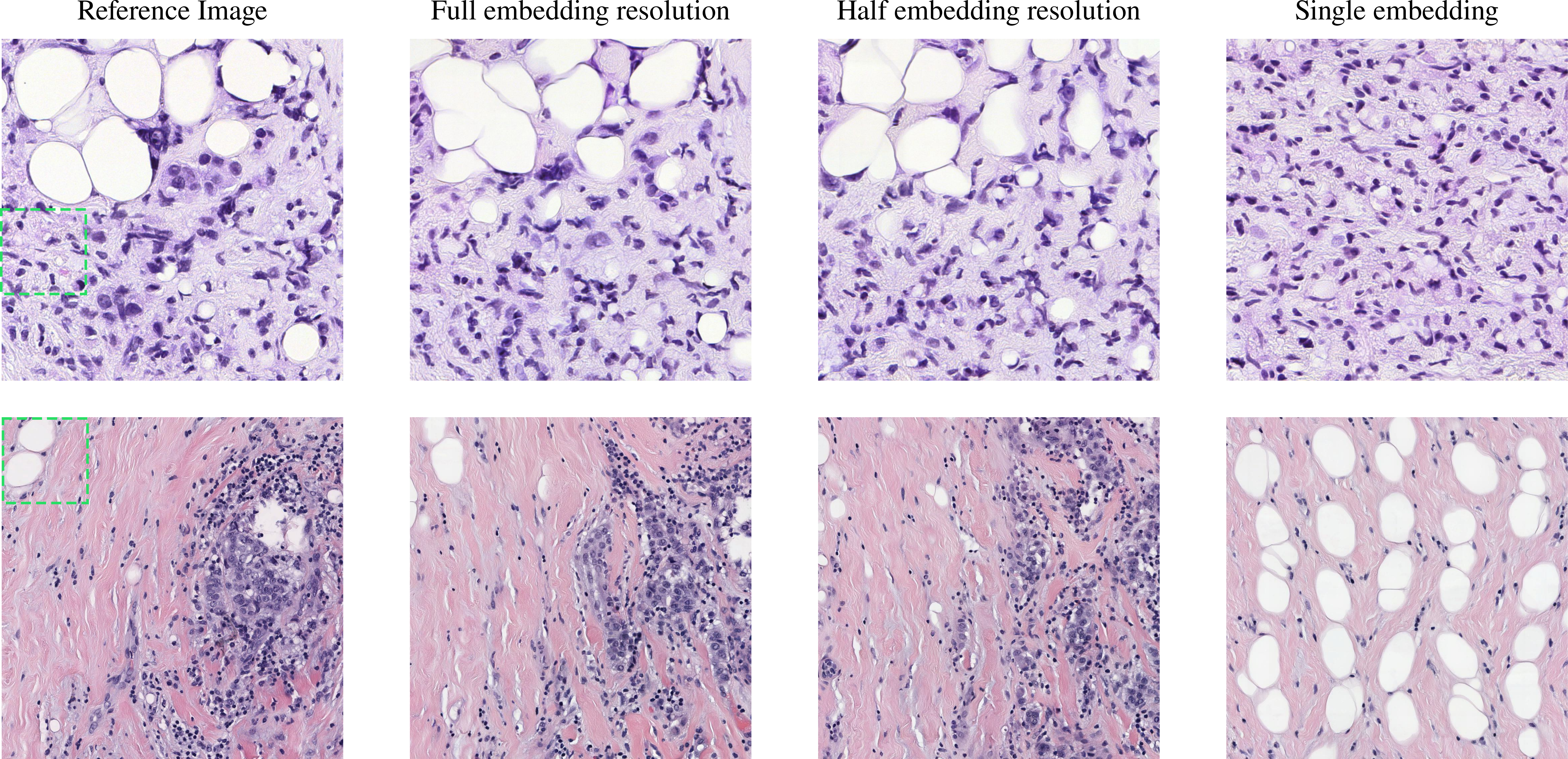}
    \caption{Using coarser conditioning results in repeated textures in the generated large image. When using a single embedding the result is equivalent to an infinitely-tiled patch. Images are at $1024 \times 1024$ pixels resolution.}
    \label{fig:embedding_res}
\end{figure}

In Fig.~\ref{fig:embedding_res} we show synthetic large images, using different embedding granularities from a reference image. When utilizing the full embedding resolution, we use the entire $4 \times 4$ embedding grid to generate a variation of the original image by interpolating to get conditioning at each $i,j$ location. At half resolution, we average the embeddings and use a $2 \times 2$ grid, leading to more repeated textures in the final image. When using a single embedding (patch indicated with a green box) the generated image is equivalent to infinitely tiling the textures from the reference patch.

\end{document}